\pdfoutput=1
\documentclass[sigconf,nonacm]{acmart} 

\usepackage[T1]{fontenc}
\usepackage[latin1]{inputenc} 
\usepackage{microtype}

\usepackage{color}
\usepackage{multirow}
\usepackage{booktabs}
\setkeys{Gin}{pagebox=artbox}
\graphicspath{{.}}

\usepackage{soul}
\setstcolor{blue}
\definecolor{violet}{rgb}{0.5,0.0,0.5}
\newsavebox\bscombox
\newcommand{\bscom}[3][]{%
	\sbox{\bscombox}{\fontsize{8}{9}\selectfont#1#2#3}
	\noindent
	\st{#2}{\selectfont
		\color{blue}#3\ifx\\#1\\\else{\fontsize{8}{9}\selectfont\color{violet}[#1]}\fi
	}
}

\begin{document}

\title{Assisted Knowledge Graph Authoring: Human-Supervised Knowledge Graph Construction from Natural Language}

\author{Marcel Gohsen}
\orcid{0000-0002-1020-6745}
\affiliation{%
  \institution{Bauhaus-Universit{\"a}t Weimar}%
  \city{Weimar}%
  \country{Germany}%
}

\author{Benno Stein}
\orcid{0000-0001-9033-2217}
\affiliation{%
  \institution{Bauhaus-Universit{\"a}t Weimar}%
  \city{Weimar}%
  \country{Germany}%
}

\renewcommand{\shortauthors}{Gohsen and Stein}


\begin{abstract}
Encyclopedic knowledge graphs, such as Wikidata, host an extensive repository of millions of knowledge statements. However, domain-specific knowledge from fields such as history, physics, or medicine is significantly underrepresented in those graphs. Although few domain-specific knowledge graphs exist (e.g., Pubmed for medicine), developing specialized retrieval applications for many domains still requires constructing knowledge graphs from scratch. To facilitate knowledge graph construction, we introduce WAKA: a Web application that allows domain experts to create knowledge graphs through the medium with which they are most familiar: natural language. 
\end{abstract}


\maketitle

\section{Motivation and Background}

Knowledge graphs are a way to encode and store real world knowledge by specifying semantic relationships (edges) between entities (nodes). These datastructures have been used effectively for various information retrieval applications like question answering \cite{huang:2019}, recommendation systems \cite{wang:2019a}, or search engines \cite{liu:2020}. Knowledge graphs are also popular for domain-specific applications in areas such as cybersecurity, education, finance, medicine, or news \cite{zou:2020}, which typically require domain-specific ontologies or knowledge.

Therefore, besides encyclopedic (e.g., Wikidata\footnote{\url{https://www.wikidata.org/}}) and commonsense (e.g., OpenCyc\footnote{\url{https://sourceforge.net/projects/opencyc/}}) knowledge graphs, there are also domain-specific graphs, for example for medicine (e.g., Pubmed \footnote{\url{https://pubmed.ncbi.nlm.nih.gov/}}) or linguistics (e.g., WordNet\footnote{\url{https://wordnet.princeton.edu/}}). However, domain-specific knowledge graphs exist only for few domains. The remaining domains require a considerable amount of effort to create knowledge graphs in order to use them in domain-specific information systems.

\begin{figure}
\includegraphics[width=.9\linewidth]{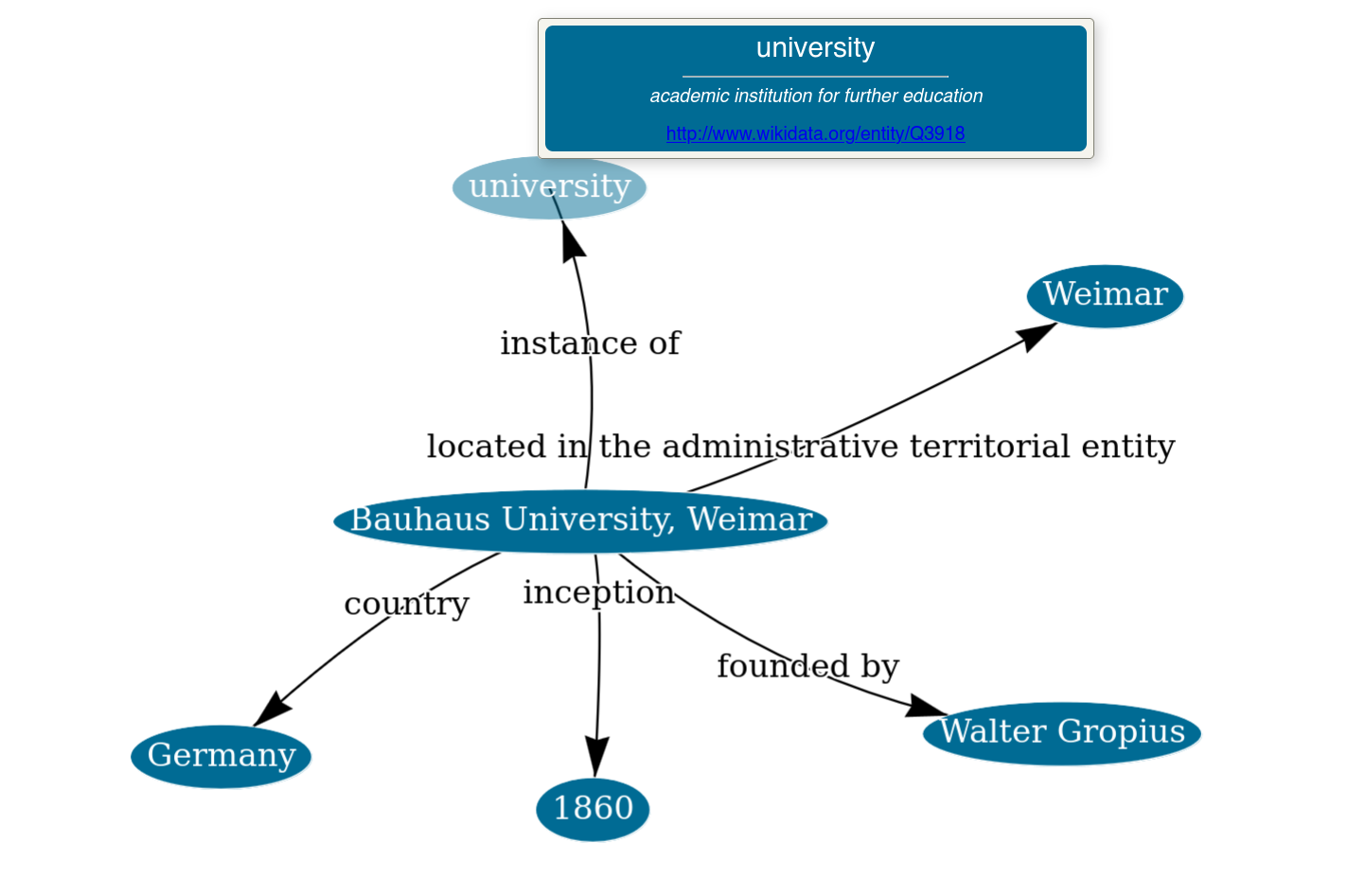}
\caption{Visualization of a knowledge graph in the WAKA frontend in which the entity \texttt{university} is highlighted.}
\label{kg-visualization}
\Description{The image shows a knowledge graph that encodes that the Bauhaus University is a university in Weimar, Germany, founded by Walter Gropius in 1860. The entity ``university'' is highlighted in the graph and a tooltip indicates that a university is an academic institution for further education and the corresponding URL in Wikidata.}
\end{figure}

The manual creation of knowledge graphs and associated ontologies from scratch is a complex process and usually requires much expertise and effort \cite{agrawal:2022}. To tackle this, automatic methods for constructing knowledge graphs from unstructured text have been realized using open information extraction \cite{martinez-rodriguez:2018}, a mixture of information extraction pipelines \cite{kertkeidkachorn:2017} or recently large language models \cite{carta:2023}. Since there was no sufficient ground truth during the development of mentioned methods, the quantitative evaluations of these methods have been limited to components of the construction algorithm (e.g. relation extraction, entity linking), but have never been extended to the entire construction algorithm. Qualitative evaluations by \citet{martinez-rodriguez:2018} and \citet{kertkeidkachorn:2017} have shown a precision of the resulting triples of 50\%, which underlines the need for human intervention in the construction process. Unfortunately, the above approaches do not provide the source code for the above knowledge graph construction approaches and therefore cannot be included in our experiments. 

With this paper, we introduce WAKA: a convenient Web application to construct and author knowledge graphs from unstructered text.\footnote{Demo video: \url{https://youtu.be/CwkW3Xwb5zg}\\ \phantom{\textsuperscript{5}}Code: \url{https://github.com/webis-de/waka}} WAKA takes advantage of the high-quality ontology and the large number of entities in Wikidata by linking entities and relations to corresponding entries in the knowledge base but also allows adding new entities and relations. The application offers an intuitive interface that allows to supervise, correct, extend, and visualize the automatically proposed knowledge graph. Finally, the authored knowledge graph can be stored in a standard data model for Semantic Web---the Resource Description Framework\footnote{\url{https://www.w3.org/RDF/}} (RDF).

\section{Authoring Interface}

The interface of WAKA consists of two  main components: a text editor and an interactive graph visualization (see Figure~\ref{kg-visualization}). A user may write or import a text into the editor from which knowledge is to be derived. Based on the text in the editor, pressing a button triggers the automatic construction of a knowledge graph (see Section~\ref{knowledge-graph-construction}), which can then be corrected and extended by the user. 

The proposed graph is shown to the user in the visualization and the corresponding entities are annotated in the text editor. For each of the entities that participate in a relation, all the mentions in the text are annotated. Since relations are not always explicitly mentioned but can be inferred from the text, we do not annotate relations in the editor. The editor and the visualization represent linked data views. Hovering over a node in the graph highlights the node and all corresponding annotations and vice versa. Since the resulting graphs can get quite large and the visualization may get cluttered, WAKA supports navigation and graph manipulation techniques such as zooming, camera movement or node dragging. 

A user can use several interactions to validate and correct the proposed knowledge graph via WAKA's interface. Hovering over an entity annotation or over a graph node reveals a label, a description, and a link to the corresponding Wikidata entity in a tooltip. Hovering over edges in the knowledge graph, displays the label, description, and Wikidata link of the associated Wikidata property. 

To correct a wrong link to a Wikidata entity or property, a user can click either on one of the corresponding annotations or on the node in the visualization. An overlay menu will be opened which shows the label and description of the currently linked entity or property. Additionally, a list of other proposed entities or properties is provided that originate from giving the mention span to the entity retrieval pipeline described in Section~\ref{entity-discovery}. Clicking on any of the proposals will establish the connection of the entity mention or relation to the Wikidata entry and the annotations and visualization will be updated accordingly. If the correct entity or property is not part of the proposals, a search box allows to freely enter a query to retrieve the correct Wikidata entry from the entity retrieval or relation retrieval pipelines, respectively. For the deletion of entities or relations a button is provided in the overlay menu. 

Entities can be added to the graph by highlighting a text span in the editor. When a text span is highlighted a button appears that opens the overlay menu to let the user select an entity from Wikidata (or leave it unlinked if the entity does not exist yet). Analogously, relationships can be added by selecting two entities and pressing a button or connecting two nodes in the visualization. 

The user can download the resulting knowledge graph by clicking on a download button in the interface. It is planned to make the download format configurable. However, WAKA currently supports to download the authored knowledge graph in N-Triples format (i.e., a plain text serialisation for RDF triples).

The knowledge graph visualization has been realized with vis-network.\footnote{\url{https://visjs.github.io/vis-network/docs/network/}} All other interface components are implemented in native Javascript without any required dependencies.
\section{Knowledge Graph Construction}
\label{knowledge-graph-construction}

\begin{figure*}
\centering
\includegraphics[width=.8\textwidth]{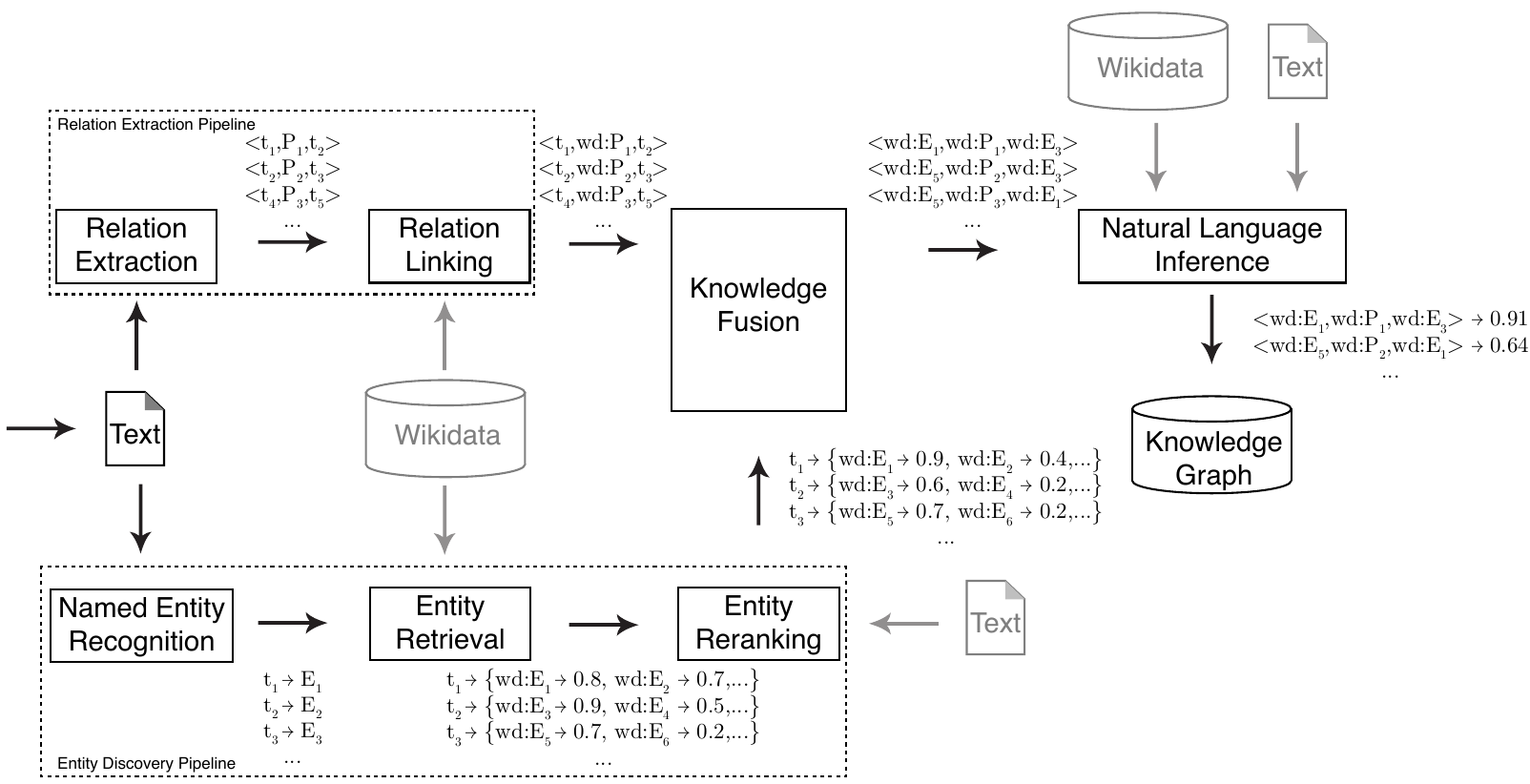}
\caption{Architecture of the automatic knowledge graph construction approach.}
\Description{The image contains a flow chart that describes the architecture of the knowledge graph construction method. Two parallel pipelines can be seen, namely the relation extraction and entity discovery pipelines. The relation extraction pipeline includes the steps of relation extraction and linking, while the entity discovery pipeline includes the steps of named entity recognition, entity search and entity reranking. The results of both pipelines are combined in the knowledge fusion step and fed into the natural language inference step.}
\label{automatic-kg-creation}
\end{figure*}

Our approach to automatically construct knowledge graphs from unstructered text consists of three main components: an entity discovery pipeline, a relation extraction pipeline, and final knowledge fusion and evaluation steps. Figure~\ref{automatic-kg-creation} gives an overview of the steps in the automatic knowledge graph construction algorithm. For efficiency reasons, the entity discovery and relation extraction pipelines are executed in parallel.

\subsection{Entity Discovery}
\label{entity-discovery}

The entity discovery pipeline aims to detect all named entities in the text, link them to entities in Wikidata and rank them according to their relevance. We calculate the relevance of an entity based on the findings of \citet{kasturia:2022}, who have found that the relevance of an entity depends equally on how well it fits a mention and a context. The set of discovered named entities forms the pool from which subjects and objects are drawn in the knowledge fusion step, rendering a high recall preferable to high precision.

\paragraph{Named Entity Recognition}
The first step in the entity discovery pipeline is to recognize named entities in the given text along with their mention spans and entity types. In addition, we extract concepts as noun phrases that were not recognized as named entities. We use established named entity recognition methods that offer a good tradeoff between effectiveness and efficiency. 

According to a comparative study of open named entity recognition frameworks \cite{schmitt:2019}, the Stanford NLP Toolkit~\cite{manning:2014} achieved the highest recall on the CoNLL 2003 dataset \cite{sang:2003}. The second-highest recall was achieved by SpaCy from which we use the en\_core\_web\_sm pipeline. Both models take features at sentence-level into account, and thus we employ the FLERT model \cite{schweter:2020} through the Flair framework \cite{akbik:2019} which considers document-level features for the named entity recognition. From all three frameworks, we employ model versions that were trained on OntoNotes 5 \cite{weischedelralph:2013}, and thus classifies found entities into an 18-class ontology. 

Based on the determined type of recognized entity, an entity must be either linked to an entry in the knowledge base or represents a literal. We distinguish between numeric literals, consisting of \texttt{PERCENT}, \texttt{MONEY}, \texttt{QUANTITY}, \texttt{CARDINAL}, and \texttt{ORDINAL} and temporal types, such as \texttt{DATE} and \texttt{TIME} from the OntoNotes ontology. Entities of the remaining entity types are linked to an entity in Wikidata.

\paragraph{Entity Retrieval}

To leave the entity disambiguation to the user, we model entity linking as a retrieval task. We build an Elasticsearch index of entities from Wikidata (i.e., subjects and objects of RDF statements). To keep Wikidata meta-information and irrelevant entities out of the index, we define the following filtering rules: 
\begin{enumerate}
\item An entity has a valid URI.
\item An entity has at least one property (i.e., outoing edge).
\item An entity is not a Wikimedia category.
\item An entity is not a Wikimedia disambiguation page.
\end{enumerate}
Following this rule set, we have collected about 109 million entities from Wikidata which is well within range of Wikidatas self-reported number of content pages of 107 million.\footnote{\url{https://www.wikidata.org/wiki/Special:Statistics}} For each entity, we store its URI, its label (i.e., the entity name displayed on its  Wikidata page), its description, and a concatenation of the label, description, and any available alternate names in English as a search key. To approximate the `commonness' of an entity for the retrieval ranking, we store the number of incoming edges for each entity. 

Each text span that represents an entity according to the named entity recognition models is queried against our entity index and ranked by BM25. However, we take into account that direct matches of the entity label are more relevant than a match of the description or alternate names. We compute the relevance score of an entity $e$ according to an entity mention $t$ as the maximum of BM25 according to the label scaled by a parameter $\alpha$ with $\alpha > 1$ and the (unscaled) BM25 according to the search key of an entity.
\begin{equation}
rel_t(e) = max\left(\alpha \cdot BM25\left(t, label\left(e\right)\right), BM25\left(t, key\left(e\right)\right)\right)
\end{equation}
Pilot experiments have shown, that $\alpha = 3$ yielded best results. In addition to an entities' relevance, we take its commonness into account. We scale the relevance of an entity by its commonness $comm(e)$. To tune down the weight of the commonness, which can reach into the millions, we take the common logarithm and add one to account for cases where the commonness is zero.
\begin{equation}
 score_t(e) = rel_t(e) \cdot log(comm(e) + 1)
\end{equation}
We retrieve at most 20 entities per mention and define a minimum score threshold of at least 20 for considered entities.

\paragraph{Entity Reranking}

The retrieval score of an entity approximates how well an entity fits to a span of a text. However, the context in which an entity is mentioned does not influence the retrieval. To remedy this, we perform a reranking that evaluates how well an entity fits semantically with the sentence it is mentioned in.

To calculate the semantic similarity between an entity and a sentence from the text, we construct a short descriptive sentence for the entity using the following template \textit{``\{label\} is a \{description\}''}. For example, if we follow the template for the named entity \texttt{Germany}, we obtain the descriptive sentence \textit{``Germany is a country in Central Europe''}. We embed the descriptive sentence and the sentence from the text with DistilRoBERTa used through the Sentence Transformer framework \cite{reimers:2019}. This model has been chosen since it provides semantically representative embeddings in a reasonable inference time. The semantic similarity is calculated by the cosine similarity between the embedding vectors and is then multiplied by the normalized retrieval score.

\subsection{Relation Extraction and Linking}

Relation extraction is the task of detecting semantic relations between entities that can be inferred from a text. In contrast to open information extraction, these relations are usually grounded in an existing ontology. The result of the relation extraction are triples representing edges in a knowledge graph. 

To solve relation extraction, we use mREBEL \cite{cabot:2023}, a state-of-the-art relation extraction  model, which is a multilingual extension to the original REBEL model \cite{cabot:2021}. Although we focus on extracting knowledge graphs from English texts, we found that mREBEL performs better and supports more relation types than the original model. The creators of mREBEL defined relation extraction as a seq2seq task and solved it by fine-tuning BART. 

A major advantage of mREBEL is that it has been trained using aligned pairs of Wikipedia abstracts and relations from Wikidata. Consequently, it is trivial to link the extracted relations to the corresponding relation in Wikidata.  

We link the extracted relations to the corresponding Wikidata relations by using the same retrieval pipeline than for the retrieval of entities (cf., Section \ref{entity-discovery}) on a separate relation index. However, no reranking is necessary, and we disambiguate retrieved Wikidata relations by choosing the best  according to its retrieval score.

\subsection{Knowledge Fusion}

The knowledge fusion components of the algorithm aims to combine knowledge triples from the relation extraction pipeline with the linked entities from the entity discovery pipeline to build Wikidata grounded RDF triples. It does so by picking subject and object entities for each extracted relation from the pool of linked entities. 

Given a ranked list of entities for each recognized mention span and a set of triples describing which entity mention spans are in relation (e.g., \texttt{<}\textit{Weimar}\texttt{, wd:country, }\textit{Germany}\texttt{>}), we replace the mention spans with their associated entity. We build a set of RDF triple candidates by applying the cartesian product out of all mentioned subject and object entities for each extracted relation.

\subsection{Natural Language Inference}

The natural language inference step ranks the RDF triple candidates by whether (1) the triple exist in Wikidata and (2) the triple can be inferred from the text. To also take the entity retrieval score into account, we first assign the mean retrieval score of subject and object entities as score for a triple candidate. If a triple candidate does exist in Wikidata, it is most likely true, and thus we boost the triples score by multiplying it with a constant factor. Prior experiments revealed that the most effective factor is three. 

To assess whether a triple can be inferred from the text is a more challenging task. We base this computation on Facebook's BART-large-MNLI model\footnote{\url{https://huggingface.co/facebook/bart-large-mnli}}---a zero-shot natural language inference (NLI) model based on BART-large \cite{lewis:2019} that was tuned on the multi-genre NLI dataset MultiNLI \cite{williams:2018}. The model predicts probabilities of whether a text is about a set of freely selectable class labels. In our case, we generate the class label of a triple by concatinating the labels of the subject, predicate, and object. To disambiguate the subject and object, we add the description in brackets. We scale the score of each triple candidate by their corresponding probability. For each extracted relation, we select the highest ranked triple candidate as the final set of triples.
\section{Evaluation}

\begin{table}[t]
\caption{Precision, Recall, and F1 of the knowledge graph construction components on the test set of the RED\textsuperscript{FM} dataset.}
\label{table-pipeline-effectiveness}
\renewcommand{\arraystretch}{1.0}
\setlength{\tabcolsep}{4pt}
\centering
\begin{tabular}{@{}l c c c@{}}
\toprule
\bf Task & \bf Precision & \bf Recall & \bf F1\\
\midrule
Named Entity Recognition & 0.067 & 0.912 & 0.121\\
Entity Retrieval  & 0.004 & 0.765 & 0.007\\
Entity Reranking  & 0.011 & 0.739 & 0.020\\
\addlinespace
Relation Extraction & 0.303 & 0.778 & 0.407 \\
Relation Linking & 0.303 & 0.778 & 0.407 \\
\addlinespace
Knowledge Fusion  & 0.147 & 0.309 & 0.180\\
Natural Lanugage Inference & 0.182 & 0.325 & 0.206\\
\bottomrule
\end{tabular}
\end{table}

To test the performance of the automatic knowledge graph construction algorithm, a ground truth of aligned texts and Wikidata grounded triples is required. There has been a lack of sufficiently large knowledge graph construction benchmark datasets. However, a new dataset for evaluating relation extraction, which also contains links to their entities and relations in Wikidata, have been released recently: the RED\textsuperscript{FM} dataset \cite{cabot:2023}. The test set of this corpus contains 446 aligned pairs of Wikipedia abstracts and sets of Wikidata-linked knowledge triples. With this dataset we compute precision, recall and F1 of all components of the construction algorithm. 

For named entity recognition, we consider a `hit' if the correct mention span is recognized. For entity retrieval and entity reranking, we compare the mention span and the Wikidata link. To evaluate relation extraction and relation linking, we compare the presence of the corresponding relation or Wikidata property, respectively. A triple is considered a hit if there is a corresponding triple with an identical subject, predicate, and object in the ground truth. To evaluate the knowledge fusion, we select the best triple by score for each set of triple candidates.

Table~\ref{table-pipeline-effectiveness} shows the macro-averaged precision, recall, and F1 values of each step in the knowledge graph construction pipeline. The entity discovery pipeline yields solid recall values which is important for the knowledge fusion. We can also see an increase in precision after reranking, without affecting recall too much. 

The relation extraction and linking pipelines also scored high recall values. However, high precision would have been desirable as this pipeline controls how many triples are returned and therefore sets an upper limit on the precision that can be achieved in the final fusion and inference steps. Consequently, the precision values for knowledge fusion and natural language inference are quite disappointing. However, the natural language inference step is able to increase both, precision and recall values of the algorithm.

In an error analysis, we found that some `errors' are actually related to the test set and not to the algorithm. With an average 2.7 triples per text with an average length of 464 characters, many of the inferrable triples are not part of the ground truth which might be one of the reasons for the low precision. The low performance of the knowledge graph construction algorithm underlines the difficulty of the problem and emphasizes the need for human-supervision.
\section{Conclusion}
\label{conclusion}

In this paper, we introduced WAKA: a Web-based application to construct and author knowledge graphs from unstructered text through the convenience of an intuitive interface. As part of WAKA, we proposed a novel approach for automatically constructing linked knowledge graphs from unstructered text. However, a quantitative evaluation of the automatic approach revealed low F1 which indicates the complexitity of the task.  The complexity of the task emphasizes the need for a method to supervise the construction and to support the correction of the resulting knowledge graphs.

\begin{acks}
This work is supported by the \grantsponsor{TMWWDG}{Th\"uringer Ministerium f\"ur Wirtschaft, Wissenschaft und Digitale Gesellschaft}{https://wirtschaft.thueringen.de} (TMWWDG) under Grant \grantnum{TMWWDG}{5575/10-5} (MetaReal).
\end{acks}

\bibliographystyle{ACM-Reference-Format}
\bibliography{chiir24-assisted-kg-authoring-lit}

\end{document}